\documentclass{article}

\usepackage{arxiv}

\usepackage[utf8]{inputenc} 
\usepackage[T1]{fontenc}    
\usepackage{hyperref}       
\usepackage{url}            
\usepackage{booktabs}       
\usepackage{amsfonts}       
\usepackage{nicefrac}       
\usepackage{microtype}      
\usepackage{lipsum}		
\usepackage{graphicx}
\usepackage[numbers]{natbib}
\usepackage{doi}
\usepackage{latexsym}
\usepackage{stmaryrd}
\usepackage{amsmath,amssymb,amsfonts}
\usepackage{multicol,lipsum}
\usepackage{subcaption}
\usepackage{amsmath,amssymb}
\usepackage{multirow}
\usepackage[linguistics]{forest}

\title{Dynamic Syntax Mapping: A New Approach to Unsupervised Syntax Parsing}

\author{ 
Buvarp Gohsh\\
University of Charleston\\
\And
Woods Ali\\
University of Charleston\\
\And
Anders Michael\\
University of Charleston\\
}

\renewcommand{\shorttitle}{Dynamic Syntax Mapping: A New Approach to Unsupervised Syntax Parsing}

\hypersetup{
pdftitle={aaa},
pdfsubject={aaa},
pdfauthor={aaa},
pdfkeywords={aaa},
}

\begin{document}
\maketitle

\begin{abstract}
The intricate hierarchical structure of syntax is fundamental to the intricate and systematic nature of human language. This study investigates the premise that language models, specifically their attention distributions, can encapsulate syntactic dependencies. We introduce "Dynamic Syntax Mapping (DSM)," an innovative approach for the agnostic induction of these structures. Our method diverges from traditional syntax models which rely on predefined annotation schemata. Instead, we focus on a core characteristic inherent in dependency relations: \textit{syntactic substitutability}. This concept refers to the interchangeability of words within the same syntactic category at either end of a dependency. By leveraging this property, we generate a collection of syntactically invariant sentences, which serve as the foundation for our parsing framework. Our findings reveal that the use of an increasing array of substitutions notably enhances parsing precision on natural language data. Specifically, in the context of long-distance subject-verb agreement, DSM exhibits a remarkable advancement over prior methodologies. Furthermore, DSM's adaptability is demonstrated through its successful application in varied parsing scenarios, underscoring its broad applicability.
\end{abstract}

\keywords{Linguistic Syntax, Unsupervised Dependency Parsing}

\section{Introduction}

The advancement of large pretrained language models (LLMs) such as BERT \citep{devlin-etal-2019-bert} has been a catalyst in enhancing performance across a spectrum of natural language processing (NLP) tasks. This progress has spurred investigations into the mechanics of natural language comprehension within these models, particularly concerning their linguistic capabilities. Building on this foundation, our study tests the hypothesis that LLMs inherently contain syntactic dependencies that can be \textit{extracted} without the need for additional parameters or external guidance.

Existing literature in the realm of syntax has delved into two key areas: (1) the examination of syntactically dependent \textit{behaviors} in language models, such as their ability to handle complex subject-verb agreement scenarios \citep{marvin-linzen-2018-targeted,FeiRJ20, gulordava-etal-2018-colorless, GoldbergBert,li-etal-2021-mrn,FeiMatchStruICML22,li2017software},\nocite{fei2020boundaries,Li00WZTJL22}  and (2) the feasibility of retrieving syntactic \textit{structures} from internal representations or mechanisms within the models \citep{hewitt-manning-2019-structural, htut-bert, limisiewicz-etal-2020-universal}. The former approach, while agnostic to specific syntactic theories, lacks the clarity offered by the explicit induction of syntactic structures, which is the focus of our work.

In an exemplary study, \citet{hewitt-manning-2019-structural} developed a probe to map word representations into a new vector space, facilitating the application of a maximum spanning tree algorithm (MST) for syntactic parsing. \nocite{Wu0LZLTJ22,shi-etal-2022-effective,FeiZRJ20,wang-etal-2022-entity} However, it remains uncertain whether this method relies exclusively on the intrinsic information of the model or introduces external knowledge through the trained probe \citep{belinkov-2022-probing}. A more unequivocal strategy involves directly employing the model's internal distributions as inputs for tree induction, without supplementary training. Prior research has explored this by utilizing attention distributions from transformer-based LMs \citep{raganato-tiedemann-2018-analysis,FeiWRLJ21, htut-bert}, resonating with observations by \citet{clark-etal-2019-bert} regarding the correlation of certain BERT attention heads with dependency relations. Nevertheless, given the vast information encapsulated in LLMs, there is no inherent guarantee that the extracted parses represent purely \textit{syntactic} structures when using these representations directly.

Our paper introduces Dynamic Syntax Mapping (DSM), an innovative methodology designed to distill syntactic information by focusing on a universally applicable property of syntactic relations, namely \textit{syntactic substitutability}. This principle recognizes that syntactic frameworks categorize words based on their mutual substitutability. Utilizing this concept, we generate a series of syntactically consistent sentences, which collectively facilitate the induction of their shared syntactic structure. Our primary objective is to determine if DSM more effectively extracts syntactic information from attention mechanisms, thereby yielding more precise induced parses. By inducing structures independently of specific annotation frameworks, DSM offers insights into how syntactic relations in a model may align with or differ from established theoretical paradigms.

Our results demonstrate that DSM significantly enhances dependency parsing accuracy. We observe a direct correlation between the number of substitutions utilized and the improvement in parsing scores. Quantitative analysis further indicates that DSM's induced parses align more closely with annotation schemas where function words assume head roles. In tests involving complex subject-verb agreement structures, DSM achieves a $>$70\% increase in recall compared to previous unsupervised parsing methods. Additionally, DSM's applicability is confirmed through its successful adaptation and performance enhancement in diverse parsing algorithms, showcasing its broad generalizability.

\section{Related work}
This study contributes to the evolving field of unsupervised syntactic parsing, a domain with rich history including the generative Dependency Model with Valence (DMV) \citep{klein-manning-2004-corpus} and advanced formulations like the Compound Probabilistic Context Free Grammar (PCFG) model \citep{kim-etal-2019-compound} for constituency parsing. The foundational concept here is that syntactic structures can be deduced by constructing a Maximum Spanning Tree (MST) based on word-pair scores within a sentence. These scores reflect the likelihood of a syntactic dependency or constituent span between the words.

Our approach, Dynamic Syntax Mapping (DSM), utilizes scores derived directly from Large Language Models (LLMs) \cite{wu2023nextgpt}, following the trajectory of prior studies that have harnessed attention distributions from transformer-based architectures \citep{NIPS2017_3f5ee243,Wu0RJL21} for scoring. In the realm of dependency parsing, noteworthy examples include \citet{raganato-tiedemann-2018-analysis,FeiGraphSynAAAI21}, who employed neural machine translation models, and \citet{htut-bert} and \citet{limisiewicz-etal-2020-universal}, \nocite{FeiLasuieNIPS22,li2017code,fei2020enriching}  who leveraged BERT. For constituency parsing, \citet{Kim2020Are} developed a methodology predicated on syntactic similarity between word pairs, derived from attention distributions. This approach aligns with \citet{clark-etal-2019-bert}'s observation of specific BERT attention heads mirroring syntactic dependencies. However, it's also noted that these heads capture a variety of linguistic relationships beyond syntax, such as coreference. DSM addresses this by focusing on syntactic substitutability, as elaborated in the subsequent section. \citet{limisiewicz-etal-2020-universal} proposed an algorithm for selecting heads relevant to syntax. We argue that this approach reaches its full potential when the distributions within a head exclusively represent syntactic information.

Complementary methodologies employ BERT's contextual embeddings for parsing. For instance, \citet{wu-etal-2020-perturbed} introduced a technique that assesses the 'impact' of masking a word on the sentence's overall representations. \nocite{fei-etal-2023-scene} Additionally, \citet{hewitt-manning-2019-structural} trained a supervised probe to project vectors into a 'syntactic' space. Another innovative direction is leveraging BERT's masked language modeling objective for calculating syntactic parsing scores, as explored by \citet{hoover-etal-2021-linguistic} and \citet{zhang-hashimoto-2021-inductive}, \nocite{FeiZLJ20} inspired by the hypothesis from \citet{futrell-etal-2019-syntactic} linking syntactic dependencies to statistical measures of mutual information.

While the application of non-parametric substitutability for syntactic parsing is novel, its conceptual roots can be traced to studies in language model interpretability. \citet{papadimitriou-etal-2022-classifying-grammatical} demonstrated that BERT systematically utilizes word-order information to distinguish syntactic roles, such as subjects and objects, even when their positions are interchanged. This phenomenon is an instance of substitution: the sentences retain their syntactic structure despite altered meanings. These findings suggest that BERT's syntax comprehension is resilient to substitutions, indicating that substitutability might be a viable constraint for syntactic parsing.

\section{Methodology}\label{sec:SemanticPerturbation}

In this section, we introduce Dynamic Syntax Mapping (DSM), a methodology that encapsulates the concept of substitutability within the induction of syntactic structures, a notion often implicitly embedded in syntactic grammars. References such as \citet{hunter-intersub} and \citet{Mel2003-kq} highlight this. Essentially, words that are intersubstitutable form syntactic \textit{categories} and underpin syntactic \textit{relations}.

\subsection{Problem statement}
Our goal is to derive a tree-structured syntactic dependency representation $t_s$ for any given sentence $s$ utilizing the mechanisms or representations inherent in an LLM. We represent the target sentence $s$, with length $n$, as:
\begin{center}
    $s := <w_{(0)},...,w_{(i)},...,w_{(n-1)}>$.
\end{center}
The edges in $t_s$ are part of a set of binary \textit{syntactic} relations $R_{synt}$, with the specifics depending on the chosen formalism. We define the relationship between two words in a sentence, $w_{(i)}$ and $w_{(j)}$, as:
\begin{center}
    $Dep_{synt}(s,i,j) \in \{0, 1\}$,
\end{center}
indicating a syntactic dependency. If $ Dep_{synt}(s,i,j) = 1$, then there is a relation $r \in R_{synt}$ such that:
\begin{center}
   $w_{(i)} \leftrightarrow{r} w_{(j)}$,
\end{center}
where $w_{(i)} \leftrightarrow{r} w_{(j)}$ signifies an \textit{undirected} relation.

To induce syntactic trees, a matrix of pairwise scores between words in a sentence is necessary. In DSM, we explore the use of attention distributions from self-attention heads as potential scores. However, this method is compatible with any scoring mechanism that calculates pairwise word scores.

We define the attention distribution for word $w_{(i)}$\footnote{BERT's tokenizer segments text into subword tokens. Following \citet{clark-etal-2019-bert}, we sum and normalize to transition from token-based to word-based attention distributions.} in sentence $s$ as:
 \begin{center}
     $Att(s, i) := [a^s_{i0},...,a^s_{ii},...,a^s_{i(n-1)}]$,
 \end{center}
where $a_{ij}$ denotes the attention weight from $w_{(i)}$ to $w_{(j)}$. The sentence's overall attention matrix, $Att(s)$, is an $n\times n$ matrix with row $i$ equal to $Att(s, i)$.

\subsection{Attention distributions}
In BERT, each word in a sentence is associated with an attention distribution across all other words. Each row $i\in [0, n)$ of $Att(s)$ corresponds to the distribution for word $w_{(i)}$.

Prior studies have utilized attention distributions to deduce syntactic trees using MST algorithms applied to a sentence's attention matrix \citep{raganato-tiedemann-2018-analysis, htut-bert}. These works hypothesize that attention scores are higher for syntactically dependent words. Therefore, a correct undirected syntactic parse can be induced as $MST(Att(s)) = t_s$, if the condition in Equation \ref{eq:bad_assumption} holds. However, we posit that this assumption is flawed, as attention distributions can encapsulate various phenomena, not exclusively \textit{syntactic} in nature. For instance, an edge in an induced tree may result from high scores due to coreference or lexical similarity rather than syntactic dependence.

\subsection{Modeling Syntactic Substitutability with DSM}
We introduce \textit{syntactic substitutability} as a core component of Dynamic Syntax Mapping (DSM), a formalism-neutral approach to extract \textit{syntactic} information from attention distributions. Syntactic grammar traditionally abstracts individual lexical items to work on broader syntactic categories. This idea is formalized in the \textit{quasi-Kunze property} discussed by \citet{Mel2003-kq}, where syntactic relations from a head word, $w_{(i)}$, to a subtree rooted at $w_{(j)}$, are defined. For a relation to be syntactically valid, it should allow for substitutions of word classes $X$ into the sentence without disrupting its syntactic integrity, as illustrated in Figure \ref{fig:SubtreeSub}.

We adapt this concept to DSM by focusing on the substitution of individual words, simplifying the approach while maintaining its theoretical rigor. We do not assume directionality in these relationships.

\noindent\textbf{Definition 1.}

\textit{Modified quasi-Kunze property for DSM}: Given words $w_{(i)}$ and $w_{(j)}$, and a relation $r \in R_{synt}$, there exists a class of words $X$ where substituting $w_{(j)}$ with any word $x \in X$ in a syntactic tree with relation $w_{(i)}\leftrightarrow[]{r} w_{(j)}$ preserves the sentence's syntactic well-formedness.

Our DSM framework posits that a syntactic relation must adhere to the \textit{modified quasi-Kunze property}. We demonstrate that this principle offers a practical route to induce dependency structures.

\subsection{Generating Sentences via Substitution}
\label{sec:generatingsents}

\begin{figure}
\centering
    \small the kids run \underline{in a park} with the ball.
    
    \small the kids run \underline{to that yard} with the ball.
    \caption{Substituting the subtree rooted at `park' (underlined) for one rooted at `yard.'}
    \label{fig:SubtreeSub}
\end{figure}

\begin{figure}
\centering
    \small just thought you 'd like to know. (\textit{Target})
    
    \small \fbox{always, simply, only} thought you 'd like to know.
    
    \small just \fbox{figured, knew, think} you 'd like to know.
    
    \small just thought you 'd \fbox{love, demand, have} to know.
    
    \small just thought you 'd like to \fbox{help, talk, stay}.
    \caption{Generating sentences using DSM for a WSJ10 dataset sample with substitutions for each position.}
    \label{fig:SemPertEx}
\end{figure}

To model the property in Definition 1, DSM employs BERT's masked language modeling to predict potential substitutions. This results in empirically valid sentence variations, as shown in Figure \ref{fig:SemPertEx}. We target open-class categories like adjectives, nouns, verbs, adverbs, prepositions, and determiners for substitution using Stanza's Universal POS tagger \citep{qi2020stanza}.

This method refines syntactic categorization beyond mere POS tagging. For instance, in Figure \ref{fig:SemPertEx}, the substitutions for `thought' are not random verbs; they're specifically chosen to fit its subcategorization requirements. Substituting `thought' with an inappropriate verb, such as `eat', would lead to grammatical errors or altered syntactic structures.

Let's denote a sentence where word $x$ replaces the word at position $j$ as $s \text{\textbackslash} (x, j)$, and the set of such sentences as $S_{sub}(s, j, X)$, defined on syntactic category $X$ from Definition 1:
\begin{equation}
    S_{sub}(s, j, X) := \{s \text{\textbackslash} (x, j) | x \in X\}.
\end{equation}

\subsection{Inducing Trees with Syntactic Relations}

The next step in DSM is applying the set of syntactically invariant sentences to extract structures from attention distributions.
For a relation $r \in R_{synt}$ satisfying Definition 1, if $Dep_{synt}(s, i, j) = 1$, it indicates a syntactic category $X$ of valid substitutions. This leads to the assumption that all sentences in $S_{sub}(s, j, X)$ share the same syntactic structure as the original, $s$.

We propose a novel method to extract syntactic structures from LLM attention distributions. Instead of applying the $MST$ algorithm on a single sentence's attention distribution, we aggregate attention matrices from a set of sentences created via substitution. The attention matrix for sentence $s$, $Att_{sub}(s)$, is derived by an aggregation function, $f$, over these matrices.

We hypothesize that averaged attention distributions across syntactically invariant sentences, more accurately reflect syntactic dependencies. Our experiments investigate whether this approach, compared to previous methods, results in more accurate syntactic parses.

\section{Experiment and Discussions}

\subsection{Experiment A}

In line with established methodologies (e.g., \citealp{hoover-etal-2021-linguistic}), we evaluate Dynamic Syntax Mapping (DSM) using two benchmark English dependency parsing datasets: (1) the Wall Street Journal (WSJ10) subsection of the Penn Treebank \citep{marcus-etal-1993-building}, annotated with Stanford Dependencies \citep{de-marneffe-etal-2006-generating} for sentences of length $\leq 10$ (389 sentences), and (2) the English section of the Parallel Universal Dependencies dataset \citep{nivre-etal-2020-universal} annotated with Universal Dependencies (EN-PUD; 1000 sentences). Additionally, Surface-Syntactic Universal Dependencies annotations \citep{gerdes-etal-2018-sud} are utilized for evaluation. Section 21 of the Penn Treebank serves as our validation set, and we further assess DSM on a challenging long-distance subject-verb agreement dataset from \citet{marvin-linzen-2018-targeted}.

Our focus is on BERT \citep{devlin-etal-2019-bert}, specifically the \texttt{bert-base-uncased} model with 12 layers and 12 self-attention heads each (110M parameters). To explore the impact of model size on generalization, we also employ \texttt{bert-large-uncased} (336M parameters).

\textbf{Experiment 1.1} aims to induce syntactic trees from averaged attention distributions across all heads at a given layer. We compare the Unlabeled Undirected Attachment Score (UUAS) of trees induced solely from the target sentence's attention distributions with those resulting from DSM application. In this experiment, DSM is implemented with $k=1$, indicating one additional sentence per word, to select an effective layer. We hypothesize DSM to be effective when syntactic information is robustly represented.

\textbf{Experiment 1.2} assesses DSM's impact with an increasing number of additional sentences ($k$) for each word, applied to the best-performing layer identified in Experiment 1.1. As DSM models syntactic substitutability through sets of sentences, we anticipate a stronger effect with more appropriate substitutions.

Both experiments use Prim's algorithm \citep{6773228} to induce non-projective, undirected, and unlabeled trees from attention matrices. This approach enables exploration of syntactic substitutability effects without assuming relation directionality. The algorithm's non-projective nature is consistently applied across all comparisons.

\subsubsection{Experiment A.1: Results and Discussion}\label{exp:1}

For \texttt{bert-base-uncased}, the largest UUAS change on the validation set between target sentence and DSM application occurs at Layer 10, which is consistent across both test sets. With \texttt{bert-large-uncased}, optimal performance is noted in Layers 17 and 18, suggesting a correlation between syntactic representation robustness and layer depth.

Our results align with previous research indicating specific layers in \texttt{base} and \texttt{large} models as most effective for syntactic information retrieval. Consistent with \citet{Kim2020Are} and \citet{tenney-etal-2019-bert}, we identify Layer 10 in \texttt{base} and Layers 17 and 18 in \texttt{large} as optimal.

Experiment A.2 extends our investigation of DSM with increased substitution numbers ($k$) for each word, utilizing Layer 10 of \texttt{base-uncased} and Layer 17 of \texttt{bert-large-uncased} for subsequent experiments.

\begin{table}[h]
\centering
\begin{tabular}{c | ccc|ccc}
\hline
\multicolumn{1}{l}{} & \multicolumn{6}{c}{UUAS (Dynamic Syntax Mapping)}\\ 
\hline
\multicolumn{1}{l}{} & \multicolumn{3}{c}{WSJ10}            & \multicolumn{3}{c}{EN-PUD}         \\ 
\hline
Layer           & \textit{T.}   & $k=1$     & $\Delta$           & \textit{T.}     & $k=1$           & $\Delta$ \\ 
\hline
6               & 57.3  & 57.3      & 0.0                & 44.8   & 44.8             & 0.0      \\
7               & 56.3  & 56.4      & 0.1                & 44.2   & 44.1             & -0.1     \\
8               & 56.0  & 56.1      & 0.1               & 43.2    & 43.2             & 0.0      \\
9               & 55.9  & 55.8      & -0.1              & 43.9    & 44.0            & 0.1      \\
10              & 55.7  & 56.8      & \textbf{1.1}               & 44.3  & 44.7              & \textbf{0.4}      \\ 
\hline
\end{tabular}%
\caption{UUAS performance comparison on WSJ10 and EN-PUD for DSM ($k=1$) against target sentence only (\textit{T.}) using \texttt{bert-base-uncased}.}
\label{tab:pud1}
\end{table}

\begin{table}[h]
\centering
\begin{tabular}{c|ccccc}
\hline
\multicolumn{1}{l}{} & \multicolumn{5}{c}{DSM Performance (\texttt{bert-base-uncased}, UUAS)}\\
\hline
      & \textit{T.}    & $k=1$   & $k=3$   & $k=5$   & $k=10$  \\
\hline
WSJ10 & 55.7 & 56.8 & 57.0 & 57.3 & \textbf{57.6} \\

EN-PUD    & 44.3 & 44.7 & 45.6 & 46.2 & \textbf{46.4}   \\
\hline
\multicolumn{1}{l}{} & \multicolumn{5}{c}{DSM Performance (\texttt{bert-large-uncased}, UUAS)}\\
\hline
WSJ10 & 56.1& 56.5& 56.7& 56.7& \textbf{57.2}\\

EN-PUD    & 45.5& 45.8& 46.2& 46.6& \textbf{47.0}\\
\hline
\end{tabular}%
\caption{Comparative results on WSJ-10 and EN-PUD for DSM with increasing substitutions ($k=1, 3, 5, 10$) using \texttt{bert-base-uncased} (Layer 10) and \texttt{bert-large-uncased} (Layer 17).}
\label{tab:increasenumber}
\end{table}

\subsubsection{Experiment A.2: Results and Discussion}\label{exp:1.2}

Table \ref{tab:increasenumber} displays the outcomes as we escalate the number of substitutions in Dynamic Syntax Mapping (DSM). Both models exhibit enhancements on WSJ-10 and EN-PUD, with a slightly more pronounced improvement for EN-PUD, which features longer sentences. The consistent rise in UUAS with the addition of more sentences underpins the effectiveness of DSM in isolating syntactic information from language models. This trend also implies that the intrinsic syntactic representations in the model withstand substitutions, allowing DSM to unravel syntactic details more proficiently than relying solely on the target sentence. An in-depth analysis of these results with the \texttt{bert-base-uncased} model is provided.

We compare DSM's performance against other methods. While DSM doesn't always reach the highest UUAS scores, its comparable outcomes and consistent improvements highlight its reliability. The following experiment delves into more intricate syntactic structures for a detailed comparison with previous methodologies.
We illustrate an example of a parse tree generated using DSM versus one derived solely from the target sentence. Some edges diverge from the UD annotation but still offer valuable syntactic insights, such as the connection from `told' to `that,' rather than directly to the main verb in the embedded clause. Such instances indicate that specific annotation choices can sometimes artificially lower UUAS scores. We next quantitatively examine this hypothesis.

\begin{table}[h]
    \centering
\begin{tabular}{lccc}
\hline
                               & \multicolumn{3}{c}{EN-PUD (UUAS)} \\ \hline
\multicolumn{1}{l|}{} & \textit{T.}  & $k=10$  & $\Delta$ \\ \hline
\multicolumn{1}{l|}{UD \citep{nivre-etal-2020-universal}}        & 44.3         & 46.4    & 2.1      \\
\multicolumn{1}{l|}{SUD \citep{gerdes-etal-2018-sud}}       & 56.0         & 59.0    & \textbf{3.0}      \\ \hline
\end{tabular}%
    \caption{Comparative UUAS performance for DSM on Universal Dependencies (UD) and Surface-Syntactic Universal Dependencies (SUD) annotations in EN-PUD (\texttt{bert-base-uncased}), contrasting target sentence only (\textit{T.}) with DSM ($k=10$).}
    \label{tab:ud-v-sud}
\end{table}

\paragraph{Discussion: DSM Parses Alignment with Syntactic Formalisms.} 
To explore the influence of annotation schemata on DSM results, we reassess the SSUD-induced EN-PUD trees using a distinct syntactic formalism: the Surface-Syntactic UD (SUD) \citep{gerdes-etal-2018-sud}. SUD primarily differs from UD in treating function words as heads of relations, as observed in our qualitative analysis.

As shown in Table \ref{tab:ud-v-sud}, DSM parses score higher in SUD (59.0 vs. 46.4 UUAS), and applying DSM in SUD results in a larger gain (+3.0pts vs. +2.1pts). Examining specific relations, such as \texttt{obl} and \texttt{ccomp} in UD (low recall) versus their SUD counterparts, \texttt{comp:obj} and \texttt{comp:obl} (high recall), further validates this observation (see \hyperref[appendix:1.2-results]{Appendix B} for complete relation-wise results).

This outcome aligns with our qualitative analysis. However, \citet{kulmizev-etal-2020-neural} reached a different conclusion when evaluating BERT's preferences for UD and SUD, using a trained probe for parse induction. This suggests that distinct methodologies may recover different types of linguistic information. Further investigation into this is earmarked for future research.

\subsection{Experiment B}

In our prior experiments, DSM demonstrated promising results on standard parsing datasets. This experiment extends our investigation to challenging long-distance subject-verb agreement constructions, as examined by \citet{marvin-linzen-2018-targeted}. These constructions test hierarchically-dependent behavior in language models, particularly in differentiating between linear proximity and syntactic dependency. We focus on whether DSM can accurately predict edges between subjects and verbs in such cases.

We examine 1000 sentences from two templates used by \citet{marvin-linzen-2018-targeted}: agreement across object relative clauses (e.g. ‘\underline{The pilot} [that the minister likes] \underline{cooks}.’) and agreement across subject relative clauses (e.g. ‘\underline{The customer} [that hates the skater] \underline{swims}.’), excluding copular verbs to control syntactic representation variations.

We evaluate DSM on these syntactic structures and compare its performance with the conditional MI method from \citet{zhang-hashimoto-2021-inductive}, which previously outperformed DSM in standard tasks.

The results, presented in Table \ref{tab:SubjectVerb}, show DSM’s enhanced edge recall, particularly notable as $k$ increases. This corroborates DSM's effectiveness from previous standard dataset experiments, with an 8.4-point increase for object relative clauses and an 8.3-point increase for subject relative clauses. DSM's performance also exceeds the conditional MI method in both clause types, highlighting its robustness. Future research could investigate how accurately predicted syntactic structures influence agreement behavior in models.

\begin{table}[h]
\centering
\begin{tabular}{ccccccc}
\hline
\multicolumn{1}{l}{} & \multicolumn{5}{c}{Object Relative Clause (recall)}                                                                                                 \\ \hline
\multicolumn{1}{c}{Method} & \textit{T.} & $k=1$ & $k=3$ & $k=5$ & $k=10$                         \\ \hline
DSM                   & 71.1                         & 71.2  & 72.4  & 75.3  & \textbf{79.5} \\
Z + H      & 8.9                          & --    & --    & --    & --                             \\ \hline
\multicolumn{1}{l}{} & \multicolumn{5}{c}{Subject Relative Clause (recall)}                                    \\ \hline
DSM                   & 54.7                         & 57.9  & 60.1  & 61.2  & \textbf{63.0} \\
Z + H         & 1.9                          & --    & --    & --    & --                                                  \\ \hline
\end{tabular}%
\caption{Recall results for subject-verb edge prediction with DSM. Comparison is made between DSM ($k=1, 3, 5, 10$) and target sentence only (\textit{T.}). Also included are average scores for conditional MI trees (Z+H) from \citet{zhang-hashimoto-2021-inductive}, based on the target sentence.}
\label{tab:SubjectVerb}
\end{table}

\subsection{Experiment C}

This experiment tests DSM's effectiveness in enhancing directed dependency parsing, utilizing a different parsing algorithm from \citet{limisiewicz-etal-2020-universal} that extracts directed dependency trees from attention distributions. We integrate DSM by replacing standard attention matrices with DSM-processed matrices in the original algorithm.

The algorithm involves selecting syntactically informative attention heads based on UD relations. This step requires gold-standard parse supervision, setting an upper limit for extracting UD-like structures from BERT. After selecting heads for both dependent-to-parent and parent-to-dependent directions, we evaluate DSM's impact on accuracy and induced syntactic trees. The approach follows the original paper’s methodology and does not incorporate POS information. We adopt their best-performing method using 1000 selection sentences. The evaluation employs unlabeled and labeled attachment scores on the EN-PUD dataset.

Table \ref{tab:ud-bert-ssud} summarizes the results. In \textbf{head selection}, DSM consistently surpasses target sentence-only performance, except for a few relations like \texttt{aux} and \texttt{amod}. For \textbf{tree induction}, DSM achieves its highest improvements in unlabeled and labeled attachment scores at $k=3$. Unlike previous experiments, increasing substitutions does not always lead to better parsing accuracy, possibly due to a limit in viable substitutions for closed categories like determiners. Future research could explore a dynamic substitution approach for further enhancements.

\begin{table}[h]
\centering
\begin{tabular}{c|ccccc}
\hline
 \multicolumn{1}{l}{}      & \multicolumn{5}{c}{Dependent-to-Parent Head Selection Accuracy}                    \\ \hline
Label  & \textit{T.} & $k=1$         & $k=3$         & $k=5$         & $\Delta$(DSM, \textit{T.}) \\ \hline
nsubj  & 63.8        & 65.8          & 67.0          & \textbf{68.2} & 4.4                  \\
obj    & 91.1        & 92.7          & \textbf{93.9} & \textbf{93.9} & 2.8                  \\
det    & 97.3        & \textbf{97.4} & 96.8          & 95.7          & 0.1                  \\
case   & 88.0        & 88.0          & \textbf{88.2} & 87.9          & 0.2                  \\ \hline
 \multicolumn{1}{l}{}     & \multicolumn{5}{c}{Tree Induction Scores}                                          \\ \hline
Metric & \textit{T.} & $k=1$         & $k=3$         & $k=5$         & $\Delta$(DSM, \textit{T.}) \\ \hline
UAS    & 52.8        & 53.7          & \textbf{54.5} & 54            & 1.7                  \\
LAS    & 22.5        & 25.6          & \textbf{26.3} & 22            & 3.8                  \\ \hline
\end{tabular}%
\caption{Accuracy in dependent-to-parent head selection and tree induction scores using DSM. Comparisons are made between DSM ($k=1, 3, 5$) and the original algorithm using only the target sentence (\textit{T.}).}
\label{tab:ud-bert-ssud}
\end{table}

\section{Conclusion and Future Work}

Our comprehensive analysis across various experimental scenarios firmly establishes the significance of incorporating syntactic substitutability into the generation of syntactic dependency structures. This study reveals that, intriguingly, attention distributions in language models, although not explicitly trained for syntactic tasks, encapsulate a substantial amount of syntactic knowledge. By applying the principle of substitutability, we have been able to harness this latent syntactic data more effectively for unsupervised parsing, thereby deepening our comprehension of how attention mechanisms learn and represent syntactic relationships. Moreover, our findings hint that earlier investigations, which focused on attention mechanisms using single sentences, might have underestimated the depth of syntactic information these models possess.
The Dynamic Syntax Mapping (DSM) methodology introduced here has shown its robustness and adaptability across different parsing frameworks and diverse datasets. Looking forward, an intriguing avenue for exploration is the application of DSM in a cross-linguistic context. Such an approach would offer valuable insights into the syntactic representation of various languages within multilingual models. The interpretable structures derived intrinsically from models through DSM not only have implications for Natural Language Processing and computational linguistics but also stand to offer a novel reservoir of data for theoretical linguistics and syntactic annotation studies. Furthermore, these findings bolster the hypothesis that syntax-like structures naturally emerge within the neural network architectures of language models. 
This research paves the way for a more nuanced understanding of the inherent capabilities of language models in representing complex syntactic phenomena and offers a methodological framework that can be leveraged for further linguistic and computational investigations.

\bibliographystyle{unsrtnat}
\bibliography{references}

\begin{thebibliography}{47}
\providecommand{\natexlab}[1]{#1}
\providecommand{\url}[1]{\texttt{#1}}
\expandafter\ifx\csname urlstyle\endcsname\relax
  \providecommand{\doi}[1]{doi: #1}\else
  \providecommand{\doi}{doi: \begingroup \urlstyle{rm}\Url}\fi

\bibitem[Devlin et~al.(2019)Devlin, Chang, Lee, and Toutanova]{devlin-etal-2019-bert}
Jacob Devlin, Ming-Wei Chang, Kenton Lee, and Kristina Toutanova.
\newblock {BERT}: Pre-training of deep bidirectional transformers for language understanding.
\newblock In \emph{Proceedings of the 2019 Conference of the North {A}merican Chapter of the Association for Computational Linguistics: Human Language Technologies, Volume 1 (Long and Short Papers)}, pages 4171--4186, Minneapolis, Minnesota, June 2019. Association for Computational Linguistics.
\newblock \doi{10.18653/v1/N19-1423}.
\newblock URL \url{https://aclanthology.org/N19-1423}.

\bibitem[Marvin and Linzen(2018)]{marvin-linzen-2018-targeted}
Rebecca Marvin and Tal Linzen.
\newblock Targeted syntactic evaluation of language models.
\newblock In \emph{Proceedings of the 2018 Conference on Empirical Methods in Natural Language Processing}, pages 1192--1202, Brussels, Belgium, October-November 2018. Association for Computational Linguistics.
\newblock \doi{10.18653/v1/D18-1151}.
\newblock URL \url{https://aclanthology.org/D18-1151}.

\bibitem[Fei et~al.(2020{\natexlab{a}})Fei, Ren, and Ji]{FeiRJ20}
Hao Fei, Yafeng Ren, and Donghong Ji.
\newblock Retrofitting structure-aware transformer language model for end tasks.
\newblock In \emph{Proceedings of the 2020 Conference on Empirical Methods in Natural Language Processing}, pages 2151--2161, 2020{\natexlab{a}}.

\bibitem[Gulordava et~al.(2018)Gulordava, Bojanowski, Grave, Linzen, and Baroni]{gulordava-etal-2018-colorless}
Kristina Gulordava, Piotr Bojanowski, Edouard Grave, Tal Linzen, and Marco Baroni.
\newblock Colorless green recurrent networks dream hierarchically.
\newblock In \emph{Proceedings of the 2018 Conference of the North {A}merican Chapter of the Association for Computational Linguistics: Human Language Technologies, Volume 1 (Long Papers)}, pages 1195--1205, New Orleans, Louisiana, June 2018. Association for Computational Linguistics.
\newblock \doi{10.18653/v1/N18-1108}.
\newblock URL \url{https://aclanthology.org/N18-1108}.

\bibitem[Goldberg(2019)]{GoldbergBert}
Yoav Goldberg.
\newblock Assessing {BERT}'s syntactic abilities.
\newblock \emph{CoRR}, abs/1901.05287, 2019.
\newblock URL \url{http://arxiv.org/abs/1901.05287}.

\bibitem[Li et~al.(2021)Li, Xu, Li, Fei, Ren, and Ji]{li-etal-2021-mrn}
Jingye Li, Kang Xu, Fei Li, Hao Fei, Yafeng Ren, and Donghong Ji.
\newblock {MRN}: A locally and globally mention-based reasoning network for document-level relation extraction.
\newblock In \emph{Findings of the Association for Computational Linguistics: ACL-IJCNLP 2021}, pages 1359--1370, 2021.

\bibitem[Fei et~al.(2022{\natexlab{a}})Fei, Wu, Ren, and Zhang]{FeiMatchStruICML22}
Hao Fei, Shengqiong Wu, Yafeng Ren, and Meishan Zhang.
\newblock Matching structure for dual learning.
\newblock In \emph{Proceedings of the International Conference on Machine Learning, {ICML}}, pages 6373--6391, 2022{\natexlab{a}}.

\bibitem[Li et~al.(2017)Li, He, Zhu, and Lyu]{li2017software}
Jian Li, Pinjia He, Jieming Zhu, and Michael~R Lyu.
\newblock Software defect prediction via convolutional neural network.
\newblock In \emph{Proceedings of the 2017 International Conference on Software Quality, Reliability and Security}, pages 318--328. IEEE, 2017.

\bibitem[Fei et~al.(2020{\natexlab{b}})Fei, Ren, and Ji]{fei2020boundaries}
Hao Fei, Yafeng Ren, and Donghong Ji.
\newblock Boundaries and edges rethinking: An end-to-end neural model for overlapping entity relation extraction.
\newblock \emph{Information Processing \& Management}, 57\penalty0 (6):\penalty0 102311, 2020{\natexlab{b}}.

\bibitem[Li et~al.(2022)Li, Fei, Liu, Wu, Zhang, Teng, Ji, and Li]{Li00WZTJL22}
Jingye Li, Hao Fei, Jiang Liu, Shengqiong Wu, Meishan Zhang, Chong Teng, Donghong Ji, and Fei Li.
\newblock Unified named entity recognition as word-word relation classification.
\newblock In \emph{Proceedings of the AAAI Conference on Artificial Intelligence}, pages 10965--10973, 2022.

\bibitem[Hewitt and Manning(2019)]{hewitt-manning-2019-structural}
John Hewitt and Christopher~D. Manning.
\newblock {A} structural probe for finding syntax in word representations.
\newblock In \emph{Proceedings of the 2019 Conference of the North {A}merican Chapter of the Association for Computational Linguistics: Human Language Technologies, Volume 1 (Long and Short Papers)}, pages 4129--4138, Minneapolis, Minnesota, June 2019. Association for Computational Linguistics.
\newblock \doi{10.18653/v1/N19-1419}.
\newblock URL \url{https://aclanthology.org/N19-1419}.

\bibitem[Htut et~al.(2019)Htut, Phang, Bordia, and Bowman]{htut-bert}
Phu~Mon Htut, Jason Phang, Shikha Bordia, and Samuel~R. Bowman.
\newblock Do attention heads in {BERT} track syntactic dependencies?
\newblock \emph{CoRR}, abs/1911.12246, 2019.
\newblock URL \url{http://arxiv.org/abs/1911.12246}.

\bibitem[Limisiewicz et~al.(2020)Limisiewicz, Mare{\v{c}}ek, and Rosa]{limisiewicz-etal-2020-universal}
Tomasz Limisiewicz, David Mare{\v{c}}ek, and Rudolf Rosa.
\newblock {U}niversal {D}ependencies {A}ccording to {BERT}: {B}oth {M}ore {S}pecific and {M}ore {G}eneral.
\newblock In \emph{Findings of the Association for Computational Linguistics: EMNLP 2020}, pages 2710--2722, Online, November 2020. Association for Computational Linguistics.
\newblock \doi{10.18653/v1/2020.findings-emnlp.245}.
\newblock URL \url{https://aclanthology.org/2020.findings-emnlp.245}.

\bibitem[Wu et~al.(2022)Wu, Fei, Li, Zhang, Liu, Teng, and Ji]{Wu0LZLTJ22}
Shengqiong Wu, Hao Fei, Fei Li, Meishan Zhang, Yijiang Liu, Chong Teng, and Donghong Ji.
\newblock Mastering the explicit opinion-role interaction: Syntax-aided neural transition system for unified opinion role labeling.
\newblock In \emph{Proceedings of the Thirty-Sixth {AAAI} Conference on Artificial Intelligence}, pages 11513--11521, 2022.

\bibitem[Shi et~al.(2022)Shi, Li, Li, Fei, and Ji]{shi-etal-2022-effective}
Wenxuan Shi, Fei Li, Jingye Li, Hao Fei, and Donghong Ji.
\newblock Effective token graph modeling using a novel labeling strategy for structured sentiment analysis.
\newblock In \emph{Proceedings of the 60th Annual Meeting of the Association for Computational Linguistics (Volume 1: Long Papers)}, pages 4232--4241, 2022.

\bibitem[Fei et~al.(2020{\natexlab{c}})Fei, Zhang, Ren, and Ji]{FeiZRJ20}
Hao Fei, Yue Zhang, Yafeng Ren, and Donghong Ji.
\newblock Latent emotion memory for multi-label emotion classification.
\newblock In \emph{Proceedings of the AAAI Conference on Artificial Intelligence}, pages 7692--7699, 2020{\natexlab{c}}.

\bibitem[Wang et~al.(2022)Wang, Li, Fei, Li, Wu, Su, Shi, Ji, and Cai]{wang-etal-2022-entity}
Fengqi Wang, Fei Li, Hao Fei, Jingye Li, Shengqiong Wu, Fangfang Su, Wenxuan Shi, Donghong Ji, and Bo~Cai.
\newblock Entity-centered cross-document relation extraction.
\newblock In \emph{Proceedings of the 2022 Conference on Empirical Methods in Natural Language Processing}, pages 9871--9881, 2022.

\bibitem[Belinkov(2022)]{belinkov-2022-probing}
Yonatan Belinkov.
\newblock Probing classifiers: Promises, shortcomings, and advances.
\newblock \emph{Computational Linguistics}, 48\penalty0 (1):\penalty0 207--219, March 2022.
\newblock \doi{10.1162/coli_a_00422}.
\newblock URL \url{https://aclanthology.org/2022.cl-1.7}.

\bibitem[Raganato and Tiedemann(2018)]{raganato-tiedemann-2018-analysis}
Alessandro Raganato and J{\"o}rg Tiedemann.
\newblock An analysis of encoder representations in transformer-based machine translation.
\newblock In \emph{Proceedings of the 2018 {EMNLP} Workshop {B}lackbox{NLP}: Analyzing and Interpreting Neural Networks for {NLP}}, pages 287--297, Brussels, Belgium, November 2018. Association for Computational Linguistics.
\newblock \doi{10.18653/v1/W18-5431}.
\newblock URL \url{https://aclanthology.org/W18-5431}.

\bibitem[Fei et~al.(2021{\natexlab{a}})Fei, Wu, Ren, Li, and Ji]{FeiWRLJ21}
Hao Fei, Shengqiong Wu, Yafeng Ren, Fei Li, and Donghong Ji.
\newblock Better combine them together! integrating syntactic constituency and dependency representations for semantic role labeling.
\newblock In \emph{Findings of the Association for Computational Linguistics: {ACL/IJCNLP} 2021}, pages 549--559, 2021{\natexlab{a}}.

\bibitem[Clark et~al.(2019)Clark, Khandelwal, Levy, and Manning]{clark-etal-2019-bert}
Kevin Clark, Urvashi Khandelwal, Omer Levy, and Christopher~D. Manning.
\newblock What does {BERT} look at? an analysis of {BERT}{'}s attention.
\newblock In \emph{Proceedings of the 2019 ACL Workshop BlackboxNLP: Analyzing and Interpreting Neural Networks for NLP}, pages 276--286, Florence, Italy, August 2019. Association for Computational Linguistics.
\newblock \doi{10.18653/v1/W19-4828}.
\newblock URL \url{https://aclanthology.org/W19-4828}.

\bibitem[Klein and Manning(2004)]{klein-manning-2004-corpus}
Dan Klein and Christopher Manning.
\newblock Corpus-based induction of syntactic structure: Models of dependency and constituency.
\newblock In \emph{Proceedings of the 42nd Annual Meeting of the Association for Computational Linguistics ({ACL}-04)}, pages 478--485, Barcelona, Spain, July 2004.
\newblock \doi{10.3115/1218955.1219016}.
\newblock URL \url{https://aclanthology.org/P04-1061}.

\bibitem[Kim et~al.(2019)Kim, Dyer, and Rush]{kim-etal-2019-compound}
Yoon Kim, Chris Dyer, and Alexander Rush.
\newblock Compound probabilistic context-free grammars for grammar induction.
\newblock In \emph{Proceedings of the 57th Annual Meeting of the Association for Computational Linguistics}, pages 2369--2385, Florence, Italy, July 2019. Association for Computational Linguistics.
\newblock \doi{10.18653/v1/P19-1228}.
\newblock URL \url{https://aclanthology.org/P19-1228}.

\bibitem[Wu et~al.(2023)Wu, Fei, Qu, Ji, and Chua]{wu2023nextgpt}
Shengqiong Wu, Hao Fei, Leigang Qu, Wei Ji, and Tat-Seng Chua.
\newblock Next-gpt: Any-to-any multimodal llm, 2023.

\bibitem[Vaswani et~al.(2017)Vaswani, Shazeer, Parmar, Uszkoreit, Jones, Gomez, Kaiser, and Polosukhin]{NIPS2017_3f5ee243}
Ashish Vaswani, Noam Shazeer, Niki Parmar, Jakob Uszkoreit, Llion Jones, Aidan~N Gomez, \L~ukasz Kaiser, and Illia Polosukhin.
\newblock Attention is all you need.
\newblock In I.~Guyon, U.~Von Luxburg, S.~Bengio, H.~Wallach, R.~Fergus, S.~Vishwanathan, and R.~Garnett, editors, \emph{Advances in Neural Information Processing Systems}, volume~30. Curran Associates, Inc., 2017.
\newblock URL \url{https://proceedings.neurips.cc/paper/2017/file/3f5ee243547dee91fbd053c1c4a845aa-Paper.pdf}.

\bibitem[Wu et~al.(2021)Wu, Fei, Ren, Ji, and Li]{Wu0RJL21}
Shengqiong Wu, Hao Fei, Yafeng Ren, Donghong Ji, and Jingye Li.
\newblock Learn from syntax: Improving pair-wise aspect and opinion terms extraction with rich syntactic knowledge.
\newblock In \emph{Proceedings of the Thirtieth International Joint Conference on Artificial Intelligence}, pages 3957--3963, 2021.

\bibitem[Fei et~al.(2021{\natexlab{b}})Fei, Li, Li, and Ji]{FeiGraphSynAAAI21}
Hao Fei, Fei Li, Bobo Li, and Donghong Ji.
\newblock Encoder-decoder based unified semantic role labeling with label-aware syntax.
\newblock In \emph{Proceedings of the AAAI Conference on Artificial Intelligence}, pages 12794--12802, 2021{\natexlab{b}}.

\bibitem[Fei et~al.(2022{\natexlab{b}})Fei, Wu, Li, Li, Li, Qin, Zhang, Zhang, and Chua]{FeiLasuieNIPS22}
Hao Fei, Shengqiong Wu, Jingye Li, Bobo Li, Fei Li, Libo Qin, Meishan Zhang, Min Zhang, and Tat-Seng Chua.
\newblock Lasuie: Unifying information extraction with latent adaptive structure-aware generative language model.
\newblock In \emph{Proceedings of the Advances in Neural Information Processing Systems, NeurIPS 2022}, pages 15460--15475, 2022{\natexlab{b}}.

\bibitem[Fei et~al.(2021{\natexlab{c}})Fei, Ren, Zhang, Ji, and Liang]{fei2020enriching}
Hao Fei, Yafeng Ren, Yue Zhang, Donghong Ji, and Xiaohui Liang.
\newblock Enriching contextualized language model from knowledge graph for biomedical information extraction.
\newblock \emph{Briefings in Bioinformatics}, 22\penalty0 (3), 2021{\natexlab{c}}.

\bibitem[Kim et~al.(2020)Kim, Choi, Edmiston, and Lee]{Kim2020Are}
Taeuk Kim, Jihun Choi, Daniel Edmiston, and {Sang-goo} Lee.
\newblock Are pre-trained language models aware of phrases? simple but strong baselines for grammar induction.
\newblock In \emph{International Conference on Learning Representations}, 2020.
\newblock URL \url{https://openreview.net/forum?id=H1xPR3NtPB}.

\bibitem[Wu et~al.(2020)Wu, Chen, Kao, and Liu]{wu-etal-2020-perturbed}
Zhiyong Wu, Yun Chen, Ben Kao, and Qun Liu.
\newblock Perturbed masking: Parameter-free probing for analyzing and interpreting {BERT}.
\newblock In \emph{Proceedings of the 58th Annual Meeting of the Association for Computational Linguistics}, pages 4166--4176, Online, July 2020. Association for Computational Linguistics.
\newblock \doi{10.18653/v1/2020.acl-main.383}.
\newblock URL \url{https://aclanthology.org/2020.acl-main.383}.

\bibitem[Fei et~al.(2023)Fei, Liu, Zhang, Zhang, and Chua]{fei-etal-2023-scene}
Hao Fei, Qian Liu, Meishan Zhang, Min Zhang, and Tat-Seng Chua.
\newblock Scene graph as pivoting: Inference-time image-free unsupervised multimodal machine translation with visual scene hallucination.
\newblock In \emph{Proceedings of the 61st Annual Meeting of the Association for Computational Linguistics (Volume 1: Long Papers)}, pages 5980--5994, 2023.

\bibitem[Hoover et~al.(2021)Hoover, Du, Sordoni, and O{'}Donnell]{hoover-etal-2021-linguistic}
Jacob~Louis Hoover, Wenyu Du, Alessandro Sordoni, and Timothy~J. O{'}Donnell.
\newblock Linguistic dependencies and statistical dependence.
\newblock In \emph{Proceedings of the 2021 Conference on Empirical Methods in Natural Language Processing}, pages 2941--2963, Online and Punta Cana, Dominican Republic, November 2021. Association for Computational Linguistics.
\newblock \doi{10.18653/v1/2021.emnlp-main.234}.
\newblock URL \url{https://aclanthology.org/2021.emnlp-main.234}.

\bibitem[Zhang and Hashimoto(2021)]{zhang-hashimoto-2021-inductive}
Tianyi Zhang and Tatsunori~B. Hashimoto.
\newblock On the inductive bias of masked language modeling: From statistical to syntactic dependencies.
\newblock In \emph{Proceedings of the 2021 Conference of the North American Chapter of the Association for Computational Linguistics: Human Language Technologies}, pages 5131--5146, Online, June 2021. Association for Computational Linguistics.
\newblock \doi{10.18653/v1/2021.naacl-main.404}.
\newblock URL \url{https://aclanthology.org/2021.naacl-main.404}.

\bibitem[Fei et~al.(2020{\natexlab{d}})Fei, Zhang, Li, and Ji]{FeiZLJ20}
Hao Fei, Meishan Zhang, Fei Li, and Donghong Ji.
\newblock Cross-lingual semantic role labeling with model transfer.
\newblock \emph{{IEEE} {ACM} Trans. Audio Speech Lang. Process.}, 28:\penalty0 2427--2437, 2020{\natexlab{d}}.

\bibitem[Futrell et~al.(2019)Futrell, Qian, Gibson, Fedorenko, and Blank]{futrell-etal-2019-syntactic}
Richard Futrell, Peng Qian, Edward Gibson, Evelina Fedorenko, and Idan Blank.
\newblock Syntactic dependencies correspond to word pairs with high mutual information.
\newblock In \emph{Proceedings of the Fifth International Conference on Dependency Linguistics (Depling, SyntaxFest 2019)}, pages 3--13, Paris, France, August 2019. Association for Computational Linguistics.
\newblock \doi{10.18653/v1/W19-7703}.
\newblock URL \url{https://aclanthology.org/W19-7703}.

\bibitem[Papadimitriou et~al.(2022)Papadimitriou, Futrell, and Mahowald]{papadimitriou-etal-2022-classifying-grammatical}
Isabel Papadimitriou, Richard Futrell, and Kyle Mahowald.
\newblock When classifying grammatical role, {BERT} doesn{'}t care about word order... except when it matters.
\newblock In \emph{Proceedings of the 60th Annual Meeting of the Association for Computational Linguistics (Volume 2: Short Papers)}, pages 636--643, Dublin, Ireland, May 2022. Association for Computational Linguistics.
\newblock \doi{10.18653/v1/2022.acl-short.71}.
\newblock URL \url{https://aclanthology.org/2022.acl-short.71}.

\bibitem[Hunter(2021)]{hunter-intersub}
Tim Hunter.
\newblock The chomsky hierarchy.
\newblock In Nicholas Allott, Terje Lohndal, and Georges Rey, editors, \emph{A Companion to Chomsky}, chapter~5, pages 74--95. John Wiley \& Sons, Ltd, 2021.
\newblock ISBN 9781119598732.
\newblock \doi{https://doi.org/10.1002/9781119598732.ch5}.
\newblock URL \url{https://onlinelibrary.wiley.com/doi/abs/10.1002/9781119598732.ch5}.

\bibitem[Mel{'\v c}uk(2009)]{Mel2003-kq}
Igor Mel{'\v c}uk.
\newblock Dependency in natural language.
\newblock In Alain Polgu{\`e}re and Igor Mel{'\v c}uk, editors, \emph{Dependency in linguistic description}, volume 111 of \emph{Studies in language companion series}. John Benjamins Pub. Co, Amsterdam; Philadelphia., 2009.

\bibitem[Qi et~al.(2020)Qi, Zhang, Zhang, Bolton, and Manning]{qi2020stanza}
Peng Qi, Yuhao Zhang, Yuhui Zhang, Jason Bolton, and Christopher~D. Manning.
\newblock Stanza: A {Python} natural language processing toolkit for many human languages.
\newblock In \emph{Proceedings of the 58th Annual Meeting of the Association for Computational Linguistics: System Demonstrations}, 2020.
\newblock URL \url{https://nlp.stanford.edu/pubs/qi2020stanza.pdf}.

\bibitem[Marcus et~al.(1993)Marcus, Santorini, and Marcinkiewicz]{marcus-etal-1993-building}
Mitchell~P. Marcus, Beatrice Santorini, and Mary~Ann Marcinkiewicz.
\newblock Building a large annotated corpus of {E}nglish: The {P}enn {T}reebank.
\newblock \emph{Computational Linguistics}, 19\penalty0 (2):\penalty0 313--330, 1993.
\newblock URL \url{https://aclanthology.org/J93-2004}.

\bibitem[de~Marneffe et~al.(2006)de~Marneffe, MacCartney, and Manning]{de-marneffe-etal-2006-generating}
Marie-Catherine de~Marneffe, Bill MacCartney, and Christopher~D. Manning.
\newblock Generating typed dependency parses from phrase structure parses.
\newblock In \emph{Proceedings of the Fifth International Conference on Language Resources and Evaluation ({LREC}{'}06)}, Genoa, Italy, May 2006. European Language Resources Association (ELRA).
\newblock URL \url{http://www.lrec-conf.org/proceedings/lrec2006/pdf/440_pdf.pdf}.

\bibitem[Nivre et~al.(2020)Nivre, de~Marneffe, Ginter, Haji{\v{c}}, Manning, Pyysalo, Schuster, Tyers, and Zeman]{nivre-etal-2020-universal}
Joakim Nivre, Marie-Catherine de~Marneffe, Filip Ginter, Jan Haji{\v{c}}, Christopher~D. Manning, Sampo Pyysalo, Sebastian Schuster, Francis Tyers, and Daniel Zeman.
\newblock {U}niversal {D}ependencies v2: An evergrowing multilingual treebank collection.
\newblock In \emph{Proceedings of the Twelfth Language Resources and Evaluation Conference}, pages 4034--4043, Marseille, France, May 2020. European Language Resources Association.
\newblock ISBN 979-10-95546-34-4.
\newblock URL \url{https://aclanthology.org/2020.lrec-1.497}.

\bibitem[Gerdes et~al.(2018)Gerdes, Guillaume, Kahane, and Perrier]{gerdes-etal-2018-sud}
Kim Gerdes, Bruno Guillaume, Sylvain Kahane, and Guy Perrier.
\newblock {SUD} or surface-syntactic {U}niversal {D}ependencies: An annotation scheme near-isomorphic to {UD}.
\newblock In \emph{Proceedings of the Second Workshop on Universal Dependencies ({UDW} 2018)}, pages 66--74, Brussels, Belgium, November 2018. Association for Computational Linguistics.
\newblock \doi{10.18653/v1/W18-6008}.
\newblock URL \url{https://aclanthology.org/W18-6008}.

\bibitem[Prim(1957)]{6773228}
R.~C. Prim.
\newblock Shortest connection networks and some generalizations.
\newblock \emph{The Bell System Technical Journal}, 36\penalty0 (6):\penalty0 1389--1401, 1957.
\newblock \doi{10.1002/j.1538-7305.1957.tb01515.x}.

\bibitem[Tenney et~al.(2019)Tenney, Das, and Pavlick]{tenney-etal-2019-bert}
Ian Tenney, Dipanjan Das, and Ellie Pavlick.
\newblock {BERT} rediscovers the classical {NLP} pipeline.
\newblock In \emph{Proceedings of the 57th Annual Meeting of the Association for Computational Linguistics}, pages 4593--4601, Florence, Italy, July 2019. Association for Computational Linguistics.
\newblock \doi{10.18653/v1/P19-1452}.
\newblock URL \url{https://aclanthology.org/P19-1452}.

\bibitem[Kulmizev et~al.(2020)Kulmizev, Ravishankar, Abdou, and Nivre]{kulmizev-etal-2020-neural}
Artur Kulmizev, Vinit Ravishankar, Mostafa Abdou, and Joakim Nivre.
\newblock Do neural language models show preferences for syntactic formalisms?
\newblock In \emph{Proceedings of the 58th Annual Meeting of the Association for Computational Linguistics}, pages 4077--4091, Online, July 2020. Association for Computational Linguistics.
\newblock \doi{10.18653/v1/2020.acl-main.375}.
\newblock URL \url{https://aclanthology.org/2020.acl-main.375}.

\end{thebibliography}

\end{document}